\documentclass{bmvc2k}

\usepackage{wrapfig,lipsum,booktabs}
\usepackage{graphicx}
\usepackage{tikz}
\usepackage{comment}
\usepackage{amsmath,amssymb}
\usepackage{color}
\usepackage{epsfig}
\usepackage{multirow}
\usepackage{enumitem}
\usepackage{wrapfig}
\usepackage{textcomp,booktabs}
\usepackage{xcolor}
\usepackage{listings}
\usepackage{algorithm}
\definecolor{ocre}{rgb}{0,0,.4}
\usepackage{caption}
\usepackage[font={small,color=ocre},figurename=Fig.,labelfont={small,bf}]{caption}
\usepackage[affil-it]{authblk}

\newcommand{\myparagraph}[1]{\vspace{2pt}\noindent{\bf{#1}}}

\title{Distilling Knowledge from Self-Supervised Teacher by Embedding Graph Alignment}

\addauthor{Yuchen Ma$^*$}{y.yuchenma@gmail.com}{1}
\addauthor{Yanbei Chen$^*$}{yanbeic@gmail.com}{1}
\addauthor{Zeynep Akata}{zeynep.akata@uni-tuebingen.de}{1,2,3}

\addinstitution{University of T\"{u}bingen, Germany}
\addinstitution{MPI for Informatics, Germany}
\addinstitution{MPI for Intelligent Systems, Germany}

\runninghead{Ma, Chen, Akata}{Distill Knowledge from Self-Supervised Teacher}

\begin{document}

\maketitle

\def\thefootnote{*}\footnotetext{Equal contribution. The work was done when Yuchen Ma was enrolled in the MSc program at the University of Heidelberg.}

\begin{abstract}
Recent advances have indicated the strengths of self-supervised pre-training for improving representation learning on downstream tasks. Existing works often utilize self-supervised pre-trained models by fine-tuning on downstream tasks. However, fine-tuning does not generalize to the case when one needs to build a customized model architecture different from the self-supervised model. In this work, we formulate a new knowledge distillation framework to transfer the knowledge from self-supervised pre-trained models to any other student network by a novel approach named Embedding Graph Alignment. Specifically, inspired by the spirit of instance discrimination in self-supervised learning, we model the instance-instance relations by a graph formulation in the feature embedding space and distill the self-supervised teacher knowledge to a student network by aligning the teacher graph and the student graph. Our distillation scheme can be flexibly applied to transfer the self-supervised knowledge to enhance representation learning on various student networks. We demonstrate that our model outperforms multiple representative knowledge distillation methods on three benchmark datasets, including CIFAR100, STL10, and TinyImageNet. Code is here: {\url{https://github.com/yccm/EGA}}.
\end{abstract}

\section{Introduction}
\label{sec:intro}

Self-supervised learning models are recently shown to be successful unsupervised learners that could greatly boost representation learning on different downstream tasks \cite{wu2018unsupervised,bachman2019learning,misra2020self,he2019momentum,chen2020simple,tian2019contrastive,tian2020makes,chen2020big}. By fine-tuning a self-supervised pre-trained model, self-supervised knowledge can often bring more advanced model performance in a wide variety of downstream tasks, such as image classification, object detection, and instance segmentation \cite{he2019momentum,chen2020exploring}. However, fine-tuning a self-supervised model could be undesirable in several aspects.
First, fine-tuning only generalizes when using the same network architecture for the downstream task, thus requiring self-supervised pre-training to be performed for each customized network. Second, fine-tuning does not allow model compression, as it is infeasible to achieve knowledge transfer to a small lightweight network.

\begin{figure}[!t]
   \begin{center}
    \includegraphics[width=0.89\linewidth]{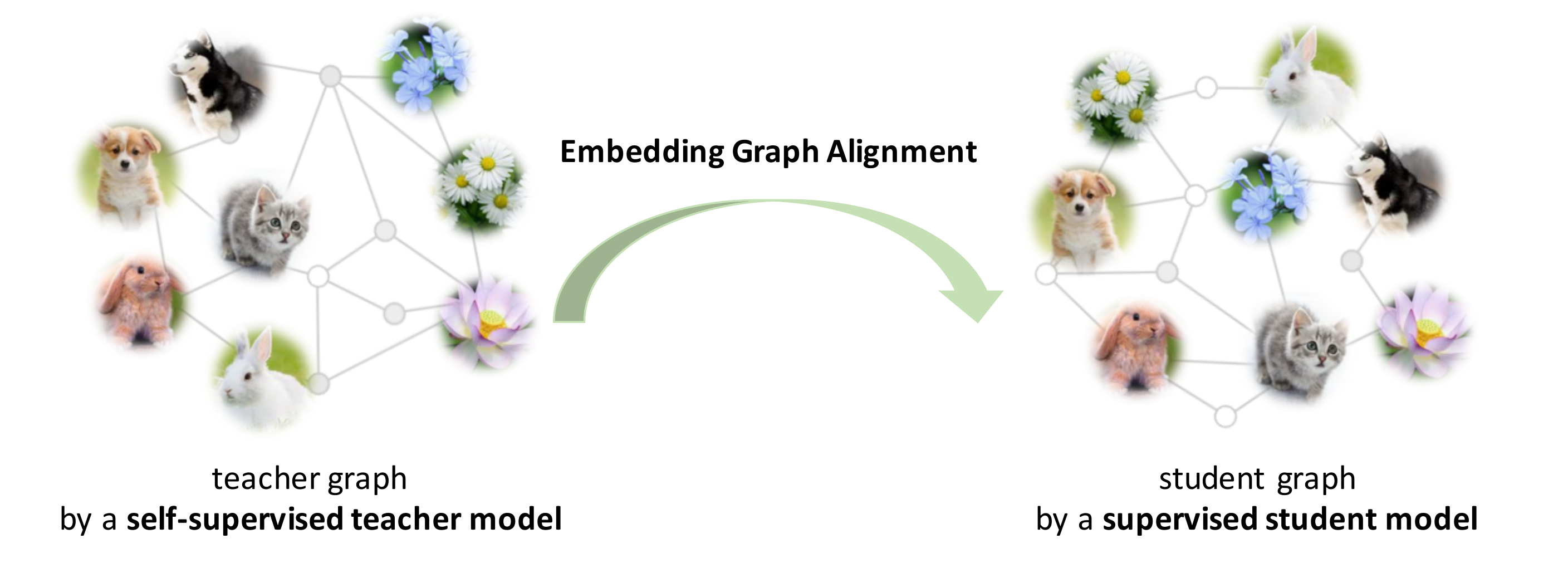}
    \end{center}
    \vskip -0.5em
    \caption{
    We consider modeling the instance-instance correlations by constructing graphs in the embedding space and transferring the graph structural knowledge from a self-supervised teacher model to a supervised student model by aligning the teacher and student graphs.
    }
    \vskip -1em
    \label{fig:teaser}
\end{figure}

In this work, to leverage the strength of self-supervised pre-trained networks, we explore the knowledge distillation paradigm \cite{hinton2015distilling,ba2014deep,bucilua2006model} rather than using fine-tuning to transfer knowledge from a self-supervised pre-trained teacher model to a small, lightweight supervised student network. To facilitate research in this regime, we first establish a knowledge distillation benchmark to test knowledge distillation upon various combinations of self-supervised teacher networks and supervised student networks. Given that the most representative self-supervised models are often trained to learn discriminative representations through instance-wise supervision signals \cite{novotny2018self,hjelm2018learning,wu2018unsupervised,bachman2019learning,misra2020self,he2019momentum,chen2020simple,tian2019contrastive,tian2020makes,chen2020big,zbontar2021barlow}, we hypothesize that these self-supervised teacher models offer a good structural information among different instances. 

To distill such fine-grained structural information learned by a self-supervised pre-trained model, we propose to model the instance-instance relationships by a graph structure in the embedding space and distill such the structural information among instances by aligning the teacher graph and the student graph -- named as Embedding Graph Alignment (EGA), as shown in Figure \ref{fig:teaser}.
Although our approach is being motivated to distill the structure of instance-instance relationships encoded in a self-supervised teacher model, our proposed model can also generalize well to distill knowledge from s supervised teacher model, therefore serving as a generic distillation scheme.

{Our contribution} is three-fold: 
(1) We propose a new knowledge distillation method called Embedding Graph Alignment (EGA), which can transfer the instance-wise structural knowledge from a self-supervised teacher model to a supervised student model by aligning the teacher graph and student graph in the embedding space. (2) We establish a comprehensive benchmark on three image classification datasets (CIFAR100, STL10, TinyImageNet) for studying knowledge distillation upon various self-supervised teacher models, comparing our method extensively to multiple state-of-the-art knowledge distillation approaches. 
(3) We demonstrate the superiority of our model under a variety of evaluation setups, including using different combinations of teacher and student models, using different training strategies in knowledge distillation, and using both self-supervised pre-trained network and supervised pre-trained network as the teacher models.

\section{Related Work}

\myparagraph{Knowledge distillation} 
is a machine learning paradigm that employs a large teacher network or an ensemble of networks to guide the learning of a small, lightweight student network 
\cite{hinton2015distilling,ba2014deep,bucilua2006model}. By mimicking the output distributions or the representations learned by a teacher model, the student network is often regularized with additional supervision signals to achieve superior model generalization, whilst such supervision is often computed as loss functions upon certain training targets. For instance, to mimic the teacher's output distributions, the first knowledge distillation (KD) approach introduces a cross-entropy loss or KL divergence computed on the {\em soft targets} from a teacher model \cite{hinton2015distilling}. To mimic the representations from a teacher model, the latter work FitNet utilizes a regression loss to align the intermediate representations between the teacher and the student networks \cite{romero2014fitnets}. The recent following works formulated more advanced and effective loss formulations for knowledge distillation \cite{zagoruyko2016paying,passalis2018learning,park2019relational,tung2019similarity,passalis2020heterogeneous}, such as aligning representations of attention maps \cite{zagoruyko2016paying}, matching the feature distributions with probabilistic models \cite{passalis2018learning}, mimicking the similarity scores of feature activations \cite{tung2019similarity}, transferring the distance-wise and angle-wise relations \cite{park2019relational}, or contrasting the representations between teacher and student \cite{tian2019contrastive}. 
Another line of following works explore to distill the knowledge learned from other datasets or data modalities \cite{gupta2016cross,gupta2016cross,aytar2016soundnet,albanie2018emotion,afouras2020asr,chen2021distilling}, such as distilling the knowledge across the audio and visual data modalities \cite{afouras2020asr,chen2021distilling}. 
In contrast to these existing works, we investigate distillation to exploit the knowledge captured from self-supervised models, and formulate a graph structure to encode the fine-grained structural instance-instance relationships learned in a self-supervised model. Our newly proposed Embedding Graph Alignment scheme particularly enables transferring the fine-grained structural semantics from the self-supervised teacher to the student network.

\myparagraph{Self-supervised learning} (SSL) is a prevalent learning paradigm that drives model training by designing a proxy protocol  to construct pseudo label supervision. By formulating unsupervised surrogate losses with free labels, SSL can learn meaningful visual representations in a fully unsupervised manner. One line of works seek for designing self-supervised pretext tasks, including predicting the expected pixel values of an output image \cite{zhang2016colorful,zhang2017split,larsson2017colorization}, solving a surrogate proxy task such as predicting {rotations} \cite{gidaris2018unsupervised}, {scale} and {tiling} \cite{noroozi2017representation}, and {patch orderings} \cite{doersch2015unsupervised,noroozi2016unsupervised,santa2018visual}. 
Another line of works propose to learn discriminative representations through instance-wise supervision signals \cite{novotny2018self,hjelm2018learning,wu2018unsupervised,bachman2019learning,misra2020self,he2019momentum,chen2020simple,tian2019contrastive,tian2020makes,chen2020big,zbontar2021barlow}, as most represented by contrastive learning methods \cite{hadsell2006dimensionality,oord2018representation,chen2020simple,he2019momentum} such as SimLR \cite{chen2020simple} and MoCo \cite{he2019momentum}; whilst the general idea of these works is to enforce instance-wise invariance towards different data augmentation applied on the same input image. 
More recently, such an instance-wise alignment scheme is further extended for training visual and language models jointly to capture the high-level semantics and learn visual representations in an unsupervised fashion \cite{radford2021learning,jia2021scaling}. In this work, for the first time, we explore the solutions to leverage the self-supervised knowledge by distillation rather than simple fine-tuning as adopted in most existing works in SSL. For this aim, we establish a new benchmark using various self-supervised teacher networks proposed in the recently advanced CLIP model \cite{radford2021learning}, and propose an effective distillation scheme that generalizes well to distill the knowledge from different types of self-supervised teacher networks.

\section{Distillation by Embedding Graph Alignment (EGA)}

\begin{figure*}[!t]
\begin{center}

\includegraphics[width=.98\textwidth]{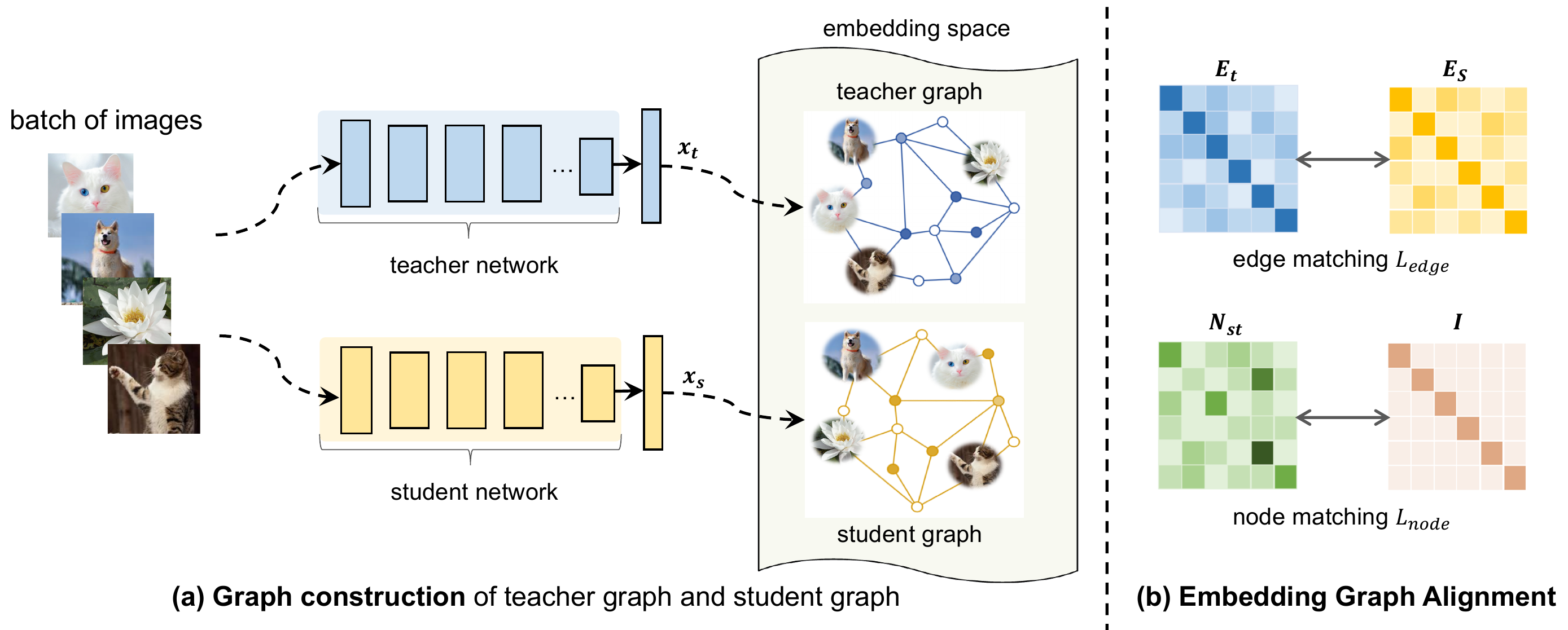}
\end{center}
\vskip -0.5em
\caption{
To distill the knowledge from the self-supervised teacher network, we first construct the teacher graph and the student graph based on a batch of embeddings to capture the {instance-instance correlation structure} in the feature embedding space (Section \ref{sec:graph}). To transfer the structural correlation information learned by the teacher network to the student network, we align the teacher graph and the student graph by jointly optimizing an edge matching constraint and a node matching constraint (Section \ref{sec:matching}). }
\label{fig:model}
\vskip -1em
\end{figure*}

Our goal is to distill the knowledge learned from a self-supervised pre-trained teacher model to enhance the visual representation learning of a supervised student model on a visual recognition task. As self-supervised learners are often trained to discriminate individual image instances \cite{wu2018unsupervised,he2019momentum,chen2020simple,chen2020exploring}, the fine-grained knowledge encoded in a self-supervised teacher network can be distilled by learning to mimic the instance-instance relationships in the feature embedding space. 
Figure \ref{fig:model} presents the model overview of our proposed Embedding Graph Alignment (EGA) approach for distilling the knowledge from a self-supervised teacher network to a supervised student network. Given a batch of $B$ image samples $\{ I_1, I_2, ..., I_B\}$, we first feed the input samples to the teacher network and the student network respectively, resulting in a batch of teacher feature embeddings $\{ \mathbf{x}_{t_1}, \mathbf{x}_{t_2}, ..., \mathbf{x}_{t_B}\}$ and a batch of student feature embeddings $\{ \mathbf{x}_{s_1}, \mathbf{x}_{s_2}, ..., \mathbf{x}_{s_B}\}$. To represent the geometry and correlations among different feature embeddings, we construct a graph $\mathcal{G}$ in the embedding space to encode the instance-instance relationships (Section \ref{sec:graph}). For distilling the knowledge, we align the teacher graph and the student graph by jointly optimizing an edge matching constraint and a node matching constraint (Section \ref{sec:matching}), which work synergistically to align the features between teacher and student networks, and transfer the instance-instance correlations from the teacher to the student. 

\subsection{Graph Construction}
\label{sec:graph}

To capture the fine-grained structural information in the embedding space, we construct a graph $\mathcal{G}=(X, E)$ to encode the structural correlations among a batch of embeddings, where $X$ refers to a set of nodes (also known as vertices), and $E$ represents a set of edges that encode the structural correlations among different instances: ${\displaystyle E\subseteq \left\{(\mathbf{x}_i,\mathbf{x}_j)\mid (\mathbf{x}_i,\mathbf{x}_j)\in X\;{\textrm {and}}\;\mathbf{x}_i\neq \mathbf{x}_j\right\}}$. In the following, we detail how we compute the node embedding and the edge to construct the graph $\mathcal{G}$ in the embedding space.

\myparagraph{Node.}
To learn the node embedding, we feed the extracted features from the teacher and student networks to individual node embedding layers (which are parameterized as linear projection layers). 
This also allows the different teacher and student networks to be deployed in our framework without checking which network needs an extra projection layer.
Let $\mathbf{f}_t \in \mathbb{R}^{D_t}$ and $\mathbf{f}_s \in \mathbb{R}^{D_s}$ denote the feature vectors extracted directly from the teacher and student networks. We pass $\mathbf{f}_t$ and $\mathbf{f}_s$ to the two node embedding layers to obtain the projected embeddings $\mathbf{x}_t \in \mathbb{R}^{D}$ and $\mathbf{x}_s \in \mathbb{R}^{D}$, where $D$ denotes the dimensionality of the embedding space shared between the teacher and the student models. The embeddings $\mathbf{x}_t$, $\mathbf{x}_s$ are utilized as the node representations in the teacher graph $\mathcal{G}_t$ and the student graph $\mathcal{G}_s$, which capture the high-level semantic information of each input image. To distill knowledge from the teacher to the student model, an intuitive solution is the align the node embeddings $\mathbf{x}_t$, $\mathbf{x}_s$ directly through certain learning constraints, which we will detail in Section \ref{sec:matching}. 

\myparagraph{Edge.} 
To capture the structural correlations among different instances, we propose to derive each edge in the graph to encode the correlation between every pair of images among the same batch based on the Pearson's correlation coefficient (PPC) -- also known as the Pearson product-moment correlation coefficient (PPMCC). 
Formally, given $n$ pairs of data points denoted as $\left\{\left(x_{1}, y_{1}\right), \ldots,\left(x_{n}, y_{n}\right)\right\}$ where $x_{i} \in \mathbb{R}^1$, 
the correlation coefficient is defined as: 
\begin{equation}
\begin{aligned}
&r_{x y} \stackrel{\text { def }}{=} \frac{\sum_{i=1}^{n}\left(x_{i}-\bar{x}\right)\left(y_{i}-\bar{y}\right)}{(n-1) s_{x} s_{y}} 
=\frac{\sum_{i=1}^{n}\left(x_{i}-\bar{x}\right)\left(y_{i}-\bar{y}\right)}{\sqrt{\sum_{i=1}^{n}\left(x_{i}-\bar{x}\right)^{2}} \sqrt{\sum_{i=1}^{n}\left(y_{i}-\bar{y}\right)^{2}}}
\end{aligned}
\label{eq:correlation}
\end{equation}
where $n$ is the sample size,
$x_{i}, y_{i}$ are the individual data points, $\bar{x}, \bar{y}$ are the mean, $s_{x}$ and $s_{y}$ are the standard deviation. 
We apply the above formula to quantify the correlation between every pair of node embedding, and construct the edge based on this correlation. Specifically, given a pair of node embeddings denoted as $\mathbf{x}$ and $\mathbf{y}$ (where $\mathbf{x} \in \mathbb{R}^D$ and $\mathbf{y} \in \mathbb{R}^D$), 

we can compute their connected edge in a graph based on Eq. \ref{eq:correlation}, as defined below. 
\begin{equation}
\begin{aligned}
e_{x,y} =\frac{\sum_{i=1}^{D}\left(\mathbf{x}_{i}-\bar{\mathbf{x}}\right)\left(\mathbf{y}_{i}-\bar{\mathbf{y}}\right)}
{\sqrt{\sum_{i=1}^{D}\left(\mathbf{x}_{i}-\bar{\mathbf{x}}\right)^{2}} \sqrt{\sum_{i=1}^{D}\left(\mathbf{y}_{i}-\bar{\mathbf{y}}\right)^{2}}}
\end{aligned}
\label{eq:edge}
\end{equation}
where $D$ is the feature dimension. 

$e_{x,y}$ is the edge that connects the two node embeddings $\mathbf{x}$, $\mathbf{y}$, and quantifies their relationship through their correlation. Based on Eq. \ref{eq:edge}, one can build the teacher graph for a batch of node embeddings $\{ \mathbf{x}_{t_1}, \mathbf{x}_{t_2}, ..., \mathbf{x}_{t_B}\}$ by computing the edge between every pairs of node embeddings, while the same procedure can be applied for building the student graph. The constructed graph, therefore, encodes the pairwise correlations among all the instances in the same batch. We will elaborate on how to exploit this structural information in the following. 

\subsection{Embedding Graph Alignment}
\label{sec:matching}

As motivated above, to distill the knowledge from a self-supervised teacher network, we aim to enforce a student network to mimic the fine-grained instance-wise correlations that are typically learned by a self-supervised learner. We encode these structural correlations by constructing the graph in the embedding space, and propose to distill such structural information by aligning the teacher graph and the student graph: named as Embedding Graph Alignment ({\bf EGA}). Below, we first detail how we construct the edge matrix and node matrix, and describe how we formulate our distillation objectives to match the edges and nodes.

\myparagraph{Edge matrix.} 
The edge matrix is formulated based on Eq. \ref{eq:edge}, which encodes all the pairwise correlations among a batch of node embeddings $X = \{ \mathbf{x}_{1}, \mathbf{x}_{2}, ..., \mathbf{x}_{B}\}$. Let $e_{i,j}$ denote the correlation between the embeddings $\mathbf{x}_{i}$ and $\mathbf{x}_{j}$ computed with Eq. \ref{eq:edge}. We can formulate an edge matrix as: 
\begin{equation}
\begin{aligned}
E (X, X) = \left(e_{i j}\right) \in \mathbb{R}^{B \times B}
\end{aligned}
\label{eq:edgematrix}
\end{equation}
where $B$ is the batch size. 
All the diagonal elements on $E$ are 1, given that its diagonal represents the self-correlations of each individual embedding. With Eq. \ref{eq:edgematrix}, we can compute the edge matrices for both the teacher model and the student model using their batch of node embeddings:  $X_t = \{ \mathbf{x}_{t_1}, \mathbf{x}_{t_2}, ..., \mathbf{x}_{t_B}\}$ and $X_s = \{ \mathbf{x}_{s_1}, \mathbf{x}_{s_2}, ..., \mathbf{x}_{s_B}\}$. Accordingly, we can write their corresponding edge matrices as $E_t = E(X_t, X_t)$ and $E_s = E(X_s, X_s)$, which encodes all the edges in two graphs.

\myparagraph{Edge matching.} 
To distill the structural correlation information captured in the teacher graph, we propose to align the edge matrices of the teacher graph and the student graph. Formally, we define an edge matching loss to match the edge information between two graphs, as written below.
\begin{equation}
\begin{aligned}
\mathcal{L}_{edge} \triangleq 
{\left\|{{E}_t - {E}_s}\right\|_{2}}
\end{aligned}
\label{eq:edgematch}
\end{equation}
where $\mathcal{L}_{edge}$ explicitly enforces the student network to distill the same structural correlations learned by the teacher network by aligning the edge matrices ${E}_t$ and ${E}_s$. Once trained with the above constraint, the student network is expected to obtain similar relative relations among different instances in the embedding space. 

\myparagraph{Node matrix.} 
Although Eq. \ref{eq:edgematch} helps to transfer the structural relations from the teacher to the student network, it does not explicitly align the individual embeddings between the teacher and the student networks. To encourage the student to mimic the representations learned by the teacher network, we propose to align their node embeddings of the same input image. Specifically, we define a node matrix where each element captures the correlation between the teacher and the student embeddings. We derive this matrix in a similar manner as our formula in Eq. \ref{eq:edgematrix}:
\begin{equation}
\begin{aligned}
N_{st} = E (X_t, X_s) 
\end{aligned}
\label{eq:nodematrix}
\end{equation}
Conceptually, the node matrix $N_{st}$ connects the teacher and student embeddings by building an edge among every pair of teacher and student embeddings to quantify the cross-correlations between the teacher and student models, 
which differs from the edge matrices that quantify the self-correlations among embeddings from the same network. 

\myparagraph{Node matching.} 
To ensure the embeddings are aligned across the teacher and student networks, we impose the following node matching loss to ensure that the correlation between the teacher and student embeddings of the same input image is high (i.e. close to 1); while the correlation between different input images is low (i.e. close to 0). For this aim, we enforce the node matrix from Eq. \ref{eq:nodematrix} to align with an identity matrix $\mathcal{I}$, as defined below. 
\begin{equation}
\begin{aligned}
\mathcal{L}_{node} \triangleq 
{\left\|{N_{st} - \mathcal{I}}\right\|_{2}}
\end{aligned}
\label{eq:nodematch}
\end{equation}
where the node matrix $N_{st}$ is encouraged to have its diagonal elements as 1 and the rest as 0, given that the diagonal encodes the correlation between the teacher and student embeddings of the same input image, and should yield high correlation values close to 1.

\myparagraph{Embedding Graph Alignment.} To align the teacher and the student graphs, we impose the embedding graph alignment loss for distillation, which contains two loss terms: the edge matching loss (Eq. \ref{eq:edgematch}) and the node matching loss (Eq. \ref{eq:nodematch}), as defined below. 

\begin{equation}
\begin{aligned}
\mathcal{L}_{EGA} = \mathcal{L}_{node} +\lambda \mathcal{L}_{edge} 
\end{aligned}
\label{eq:gm}
\end{equation}
where $\lambda$ is a hyperparameter to balance two loss terms.

\subsection{Model Optimization}
\label{sec:opt}

\myparagraph{Overall objective.} 
Our learning objective includes a standard task objective and an embedding graph alignment loss $\lambda_{EGA}$ for knowledge distillation. We consider image classification in this work and the task objective is a standard cross-entropy loss $\mathcal{L}_{CE}$. The overall objective can be written as: 

\begin{equation}
\begin{aligned}
\mathcal{L}  = \mathcal{L}_{ce} + \lambda_{EGA} \mathcal{L}_{EGA}
\end{aligned}
\label{eq:final}
\end{equation}
Given that the teacher network is a self-supervised pre-trained model and it is not optimized for the downstream target task, we consider two standard distillation strategies to train the teacher and student networks: (1) mutual learning \cite{zhang2018deep}, which optimizes the teacher with the target objective $\mathcal{L}_{ce}$ and the student with the full objective (Eq. \ref{eq:final}) {\em simultaneously}; (2) sequential learning \cite{hinton2015distilling}, which optimizes the teacher and student networks {\em sequentially} by first training the teacher with the task objective $\mathcal{L}_{ce}$ and then training the student with the full objective (Eq. \ref{eq:final}). For both strategies, the teacher's backbone is frozen and only its new added layers are trained. In our experiments, we find the two strategies perform similarly. We provide algorithm overviews on these strategies in supplementary.

\section{Experiments}

\subsection{Experimental Setup}
\label{sec:setup}

\myparagraph{Datasets.}
We evaluate on three image
classification datasets. 
{\bf CIFAR100}
consists of 60,000 32 × 32 colour images, with 50,000 images for training and 10,000 images for testing. It has 100 classes and each class contains 600 images.
{\bf STL-10} 
contains a training set of 5,000 labeled images from 10 classes and 100,000 unlabeled images, and a test set of 8K images. It has 10 classes and the image size is 96x96. We only use the labeled data for training. 
{\bf TinyImageNet} 
contains 100,000 colored images of 200 classes, and each class contains 500/50 training/validation images from ImageNet. The image sizes are downsized to 64x64.

\myparagraph{Implementation details.} 
We use PyTorch for our experiments. 
We test different types of student networks, including resnet8x4, shuffleNetV1, and VGG13. All the extracted image features from the student network are passed through a node embedding layer (Section \ref{sec:graph}), which maps each feature vector to a 256-D embedding by a linear projection. 
We also test different types of teacher networks, including self-supervised models and supervised models. Our self-supervised models include ViT-B/32, ViT-B/16 and RN101 \cite{radford2021learning}; while our supervised models include RN50, RN101 and WRN-40. Similarly, all the features from the teacher network are passed through a node embedding layer to get a 256-D embedding. For data augmentation, we apply standard strategies such as random cropping, random horizontal flipping, and normalization. For the hyperparameters, we set $\lambda$ to 0.3 and $\lambda_{EGA}$ as 0.8. For model training, we adopt the SGD optimizer, with an initial learning rate of 0.05, and decayed by 0.1 every 30 epochs after the first 150 epochs. The model is trained for 240 epochs with a batch size of 64. For distillation, we consider two training strategies: mutual learning and sequential learning as mentioned in \ref{sec:opt}. 

\subsection{Comparing to the State-of-the-Art}
\label{sec:sota}

\myparagraph{Compared methods.} 
We compare our method to multiple representative state-of-the-art distillation methods. 
To ensure fair comparisons, we adopt the same network architectures, the same optimization scheme and the training strategies for all the methods, while using the customized distillation objectives for different methods. 

we compare against the following models: 
{\bf KD} \cite{hinton2015distilling}: the first distillation method, using soft targets from the teacher to guide the student by KL divergence. {\bf FitNet} \cite{romero2014fitnets}: a popular technique that aligns the features between the teacher and student networks by regression. {\bf PKT} \cite{passalis2018learning}: a probabilistic distillation model that matches the probabilistic distributions between the student and teacher by a divergence metric. {\bf RKD} \cite{park2019relational}: 
a relational distillation technique that transfers the distance-wise and angle-wise relations among features from the teacher to student model.{\bf IRG} \cite{liu2019knowledge}:
a distillation approach that models feature space transformation across layers and minimizes its MSE loss.
{\bf CRD} \cite{tian2019contrastive}:
a contrastive distillation method that transfers knowledge by aligning the teacher and student's representations through instance-wise contrastive learning. {\bf CCL} \cite{chen2021distilling}: 
a recent distillation method to transfer knowledge across heterogeneous networks using a noise contrastive loss to align teacher and student's representations and a JSD loss to align the model predictions

\myparagraph{\underline{Discussion.}} Among all the methods above, RKD and IRG are most related to our model as these are motivated to transfer the relational knowledge learned from the teacher to the student. However, our method EGA differs from other relational distillation approaches in multiple aspects: {\bf (1)} We're the first to explore distillation from self-supervised teacher models and establish a new benchmark (Table1\&2), while showing better performance on standard distillation using supervised teacher models (Table3). {\bf (2)} We formulate graphs with node embedding layers, Pearson’s correlation coefficients (PPC, Eq.1,2), and match graphs by enforcing cross-graph node matrix to be identity and aligning edge matrices with Frobenius norm (Eq.3-6), which greatly differs from these two methods: RKD computes relations as distances between pairs, angles among triplet and aligns relations; IRG models feature space transformation across layers and minimizes its MSE loss. {\bf (3)} Our method shows 
more simplicity in implementation and achieves superior performance.

{
\setlength{\tabcolsep}{2pt}
\renewcommand{\arraystretch}{1.1}
\begin{table}[!t]
    \centering 
	\scalebox{0.7}{
	\begin{tabular}{l|c|c|c|c|c|c}
	\hline \multirow{2}{10pt}{Method} 
		& \multicolumn{3}{c|}{Same student different teacher} 
		& \multicolumn{3}{c}{Same teacher different student} \\ 
		\cline{2-4}	\cline{5-7}   & ViT-B/32 & ViT-B/16 & RN101 &Resnet8x4 & ShuffleNetV1 & VGG13
	\\	\hline	\hline

		KD\cite{hinton2015distilling} & 
		71.55 & 71.99 & 64.77 & 71.55 & 72.90 & 75.20
		\\ 
        FitNet\cite{romero2014fitnets} & 
		  73.93 & 74.13 & 74.14 & 73.93 & nan & 75.56
		\\

		PKT \cite{passalis2018learning} & 
		  73.86 & 73.55 & 72.21 & 73.86 & 75.31 & 75.55 
		\\
		RKD\cite{park2019relational} & 
         73.34 & 73.42 & 73.7 & 73.34 & 73.93 & 76.41   
		  \\	
		NCE \cite{chen2021distilling} & 
		 74.30 & 74.41 & 73.69 & 74.30 & 73.99 & 76.42
		\\
        IRG \cite{liu2019knowledge} & 
        75.11 & 74.72 & 74.17 & 75.11 & 74.79 & 75.98
		\\
        CRD\cite{tian2019contrastive} & 
		 75.73 & 75.68 & 75.13 & 75.73 & 75.54 & 76.83
	    \\

		CCL\cite{chen2021distilling} &
		75.91 & 76.13 & 75.08 & 75.91 & 76.14 & \bf 77.68
		\\
		\hline\hline
		\bf EGA &
		\bf 76.65 & \bf 76.30 & \bf 75.41 & \bf 76.65 & \bf 76.24 & 77.59 \\

		\hline

	\end{tabular}
	}
    \vskip 0.5em
	\caption{
	The teacher and student are trained {\bf \em simultaneously} \cite{zhang2018deep}. 
	Best results are {\bf bold}.
	} 
	\label{tab:network1}
 	\vskip -0.5em
\end{table}
}
{
\setlength{\tabcolsep}{2pt}
\renewcommand{\arraystretch}{1.1}
\begin{table*}[!t]
    \centering 
	\scalebox{0.7}{
	\begin{tabular}{l|c|c|c|c|c|c}
	\hline \multirow{2}{10pt}{Method} 
		& \multicolumn{3}{c|}{Same student different teacher} 
		& \multicolumn{3}{c}{Same teacher different student} \\ 
		\cline{2-4}	\cline{5-7}   & ViT-B/32 & ViT-B/16 & RN101 &Resnet8x4 & ShuffleNetV1 & VGG13
	\\	\hline	\hline
		RKD\cite{park2019relational} & 
         73.36 &  72.43 & 73.92 &73.36 & 72.62 & 73.26 
		  \\
		CRD\cite{tian2019contrastive} \quad & 
		 75.51 & 73.38 & 74.85  & 75.51 & 74.87 &77.41
		\\
		CCL\cite{chen2021distilling} \quad & 
		 75.98 & 39.56 & 74.22 & 75.98 & 76.05 &77.54 
        \\
		\hline\hline
		\bf EGA \quad & 
		\bf  76.11 & 
		\bf 74.02 &
		\bf  75.22 &
		\bf  76.11 &
		\bf  76.74 &
		\bf  77.76 \\
		\hline
	\end{tabular}
	}
    \vskip 0.5em
	\caption{
	The teacher and student 
	are trained {\bf \em sequentially} \cite{zhang2018deep}. 
	Best results are {\bf bold}.
	} 
	\label{tab:network2}
	\vskip -1em
\end{table*}
}
\begin{table}[!t]
\begin{minipage}{.45\textwidth} %
\resizebox{0.95\textwidth}{!}{
	\begin{tabular}{l|c|c|c}
	\hline
	
	\multirow{2}{10pt}{Method} 
		& \multicolumn{3}{c}{Same student different teacher}\\
		
		\cline{2-4} &  RN101 & RN50 & WRN-40 
	\\	\hline	\hline
		KD\cite{hinton2015distilling}  & 
          74.69 & 74.82 & 74.77
        \\ 
        FitNet\cite{romero2014fitnets} & 
		  58.73 & 76.27&75.58
		\\
		PKT\cite{passalis2018learning} & 
		  74.44 & 75.69&75.30
		\\
		RKD \cite{park2019relational} & 
         72.45 & 72.25&72.48
		  \\
		NCE\cite{chen2021distilling} & 
		 73.62 & 74.35& 72.90
		\\
		CRD \cite{tian2019contrastive}& 
		 75.52 & 75.50& 75.84
		\\
		CCL\cite{chen2021distilling} &
		75.56 & 75.53& 75.33
		\\
		\hline \hline
		\bf EGA & 
		\bf 75.77 & 
		\bf  76.36 
		& \bf75.97
		\\
		\hline

	\end{tabular}}
	\vskip 0.5em
\caption{Top 1 accuracy(\%) of student network on CIFAR100. 
	The teacher and student are trained 
	{\bf \em simultaneously} \cite{zhang2018deep}. 
}
\label{tab:sup}
\end{minipage}%
\hspace{10pt}
\begin{minipage}{.47\textwidth} %
\resizebox{0.95\textwidth}{!}{
\begin{tabular}{l|c|c|c}
	\hline
        Method & CIFAR100 & STL-10 & TinyImageNet \\ 
        \hline \hline
		KD \cite{hinton2015distilling} & 
		71.55 & 84.35 & 54.68
		\\
        FitNet \cite{romero2014fitnets} & 
		  76.04 & 84.15 & 59.97
		\\
        PKT \cite{passalis2018learning} & 
		  72.51 & 82.37 & 58.34
		\\
		
		RKD \cite{park2019relational} & 
         73.34 & 83.13 & 58.15 
		  \\
		NCE \cite{chen2021distilling} & 
		 74.30 & 83.96 & 58.93
		\\
		CRD \cite{tian2019contrastive} & 
		 75.73 &  82.40 & 60.34 
		\\

		CCL \cite{chen2021distilling} & 
		 75.91 & 84.01 & 60.84
		 \\

		\hline \hline
		\bf EGA & 
		\bf 76.65 & 84.15 & 60.61
		\\
	\bf EGA + KD & 
		76.49 & \bf 84.36 & \bf 61.24 \\
	\hline
	\end{tabular}}
	\vskip 0.5em
\caption{Top 1 accuracy (\%) of student networks on CIFAR100, STL10, TinyImageNet. 
	Teacher: self-supervised ViT-32 \cite{radford2021learning}. 
	Student: resnet8x4. 
	The teacher and student are trained 
	{\bf \em simultaneously} \cite{zhang2018deep}. 
	}
\label{tab:data1}
\end{minipage}
\vskip -1em
\end{table}

\myparagraph{Evaluation on different network architectures.} 
To evaluate how our model performs using different combinations of teacher and student networks, we deploy distillation using (1) the same student and different teacher, and (2) the same teacher and different student. 

Table \ref{tab:network1} shows our results on CIFAR100 with various teacher-student combinations. As shown, our EGA offers superior model performance in four out of five cases as compared to the best competitor CCL. 
When using the teacher and student as ``ViT-B/32 + resnet8x4'', ``ViT-B/16 + resnet8x4'', ``RN101 + resnet8x4'', and ``ViT-B/32 + shuffleNetV1'', EGA obtains an accuracy of 76.65\%, 76.30\%, 75.41\%, 76.24\%, which are much higher than 75.91\%, 76.13\%, 75.08\%, 76.14\%  by CCL. When using the teacher and student as  and ``ViT-B/32 + VGG13'', EGA achieves very similar performance as CCL, obtaining an accuracy of 77.59\% vs 77.68\% by CCL which combines NCE and JSD for distillation. 

In Table \ref{tab:network2}, we evaluate the sequential learning strategy of teacher and student in distillation. 
When comparing EGA to the top competitors, we find that EGA obtains the best overall performance among different combinations of teacher and student networks. In particular, EGA surpasses the best competitor CCL in five out of five cases, obtaining an accuracy of 76.11\%, 74.02\%, 75.22\%, 76.74\%, 77.76\% 
when using the teacher and student as ``ViT-B/32+resnet8x4'', ``ViT-B/16+resnet8x4'', ``RN101+resnet8x4'', ``ViT-B/32+shuffleNetV1'', and ``ViT-B/32+VGG13''. Overall, our results in Table \ref{tab:network1} and Table \ref{tab:network2} indicate that EGA achieves strong model generalization when using different combinations of teacher and student networks, under two different training strategies for distillation. These results collectively show that EGA is capable of distilling self-supervised knowledge in an effective way.

\myparagraph{Evaluation on distilling supervised knowledge.}
We focus on distilling self-supervised knowledge in this work, given that self-supervised models are cost-effective -- they require no labels but learn good features. However, our model can also generalize to distilling knowledge from self-supervised networks. We evaluate this aspect in Table \ref{tab:sup}. As shown, our model EGA achieves the best model performance among all the competitors in all the comparisons, offering an accuracy of 75.77\%, 76.36\%, 75.97\% vs 75.56\%, 75.53\%, 75.33\% by the competitor CCL when using RN101, RN50, WRN-40 as the supervised teacher network. These results indicate that our EGA also works well in the scenario where the teacher networks are trained in a supervised manner.

\myparagraph{Evaluation on CIFAR100, STL-10 and TinyImageNet.} 
Table \ref{tab:data1} shows our results on three image classification datasets, which transfer the self-supervised knowledge learned from the CLIP model \cite{radford2021learning} to a lightweight student network by training the two networks simultaneously. As Table \ref{tab:data1} shows, our model GM obtains the overall superior model performance compared to other strong competitors.   
When comparing EGA+KD with the recent best competitor CCL which combines two losses (NCE+JSD), we find that EGA+KD outperforms CCL substantially, obtaining an accuracy of 76.49\%, 84.36\%, 61.24\% on the three datasets vs 75.91\%, 84.01\%, 60.84\% by CCL.

In Table \ref{tab:data2}, we evaluate another sequential learning strategy on the three datasets. By sequential learning, we first pre-train the teacher model on the downstream task and keep its weights frozen during distillation. As can be seen, when comparing EGA to multiple top competitors, the superiority of EGA remains the same. 
Overall, our results in Table \ref{tab:data1} and Table \ref{tab:data2} show that our model EGA works well for distilling self-supervised knowledge to boost the classification performance on various downstream image datasets. Moreover, EGA also works well under different distillation strategies, i.e., training the teacher and student simultaneously or sequentially.  
These results show the superior generalizability of EGA for distilling self-supervised knowledge in different scenarios. 

\begin{wraptable}{r}{5.5cm}
    \centering 
	\scalebox{0.7}{
	\begin{tabular}{l|c|c|c}
	  
        \hline 
	     Method & CIFAR 100 & STL-10 & TinyImageNet \\
		 \hline \hline 
		RKD \cite{park2019relational}& 
		73.36 & 82.67 & 58.32
		\\ 
		CRD \cite{tian2019contrastive}& 
		75.51 & 78.76 & 60.82 
		\\ 
        CCL\cite{chen2021distilling} &
        75.98 & 80.41 & 61.24 
		\\
		\hline \hline
		\bf EGA & 
		\bf 76.11 & \bf 83.01 & \bf 61.85
		\\
		\hline 
	\end{tabular}
	}
    \vskip 0.5em
	\caption{
	Top 1 accuracy (\%) of student networks on CIFAR100, STL10, TinyImageNet. 
	Teacher: self-supervised ViT-32 \cite{radford2021learning}. 
	Student: resnet8x4. 
    The teacher and student are trained 
	{\bf \em sequentially} \cite{hinton2015distilling}. 
	Best results are highlighted in {\bf bold}. 
	} 
	\label{tab:data2}
\end{wraptable} 

\begin{figure}[!t]
    \centering
     \includegraphics[width=0.7\linewidth]{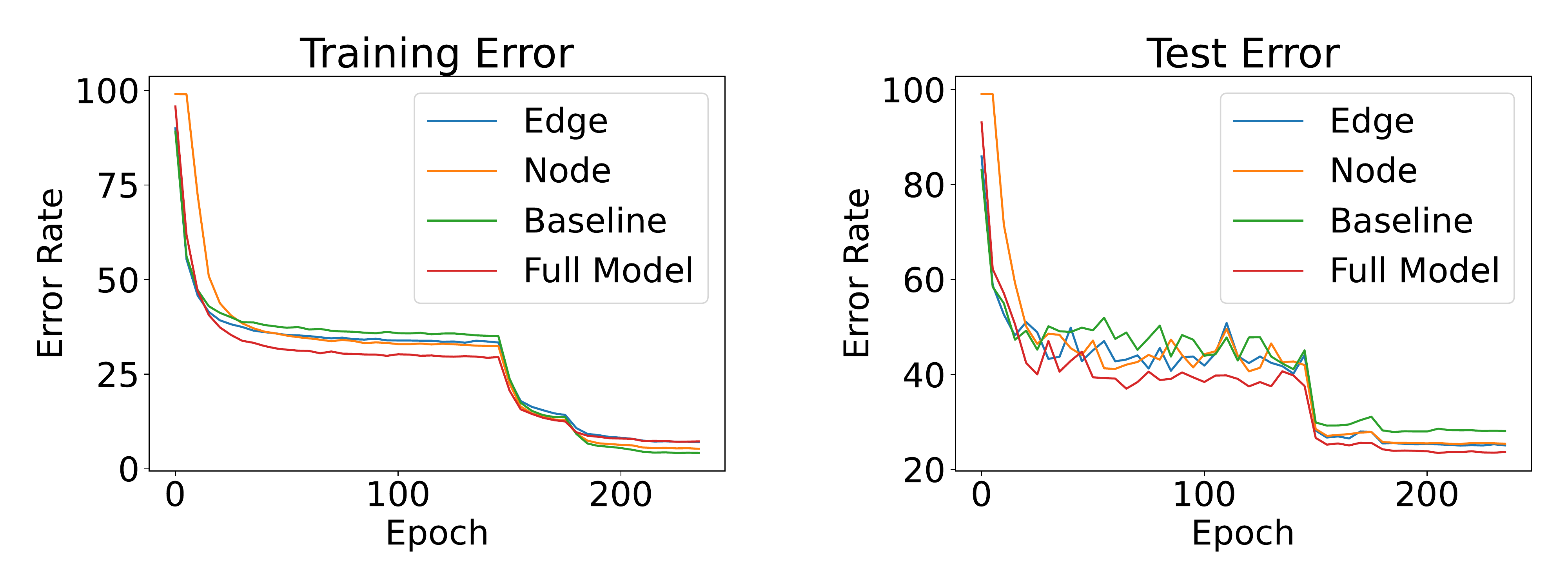}
     \vskip 0.5em
    \caption{The learning curves of baseline, EGA w edge matching loss only, EGA w node matching loss only, EGA on CIFAR100.}
    \label{fig:curve}
    \vskip -1em
\end{figure}

\myparagraph{Analyzing the learning dynamics.} 
Figure \ref{fig:curve} shows the learning curves of different models on CIFAR100. We compare the baseline model (only $\mathcal{L}_{ce}$), edge matching loss model ($\mathcal{L}_{ce}+\lambda_{edge}\mathcal{L}_{edge}$), node matching loss model ($\mathcal{L}_{ce}+\lambda_{node}\mathcal{L}_{node}$) and our full EGA model. It is interesting to observe that our EGA is less overfitting on the training set compared to other models, especially the baseline. We find that EGA consistently performs best during training on the test set, obtaining the lowest error rate. Besides, when adding either $\mathcal{L}_{edge}$ or $\mathcal{L}_{node}$ with the $\mathcal{L}_{ce}$, the model outperforms the baseline, which indicates that these two loss terms are indeed effective. Figure \ref{fig:curve} again proves that our full model formulation EGA performs and generalizes much better than the EGA variants with a single loss term.

\vskip -2em
\section{Conclusion}
In this work, we introduced a new Embedding Graph Alignment (EGA) method for distilling self-supervised knowledge. To capture the fine-grained instance-wise correlations learned by the teacher network, we construct graphs to encode the structural information in the embedding space, and perform knowledge distillation by aligning the teacher graph and the student graph. We evaluated our EGA extensively on multiple image classification datasets in comparison to a wide variety of knowledge distillation methods. We show that EGA generalizes well on different datasets, under different combinations of teacher and student networks, and performs robustly when using different training strategies for distillation. Our results collectively demonstrate the superiority of EGA as a generic knowledge distillation technique.

\myparagraph{Acknowledgement.} This work has been partially funded by the ERC (853489-DEXIM) and the DFG (2064/1–Project number 390727645).

\section*{\Large Supplementary Material}
\setcounter{figure}{0}
\setcounter{section}{0}
\setcounter{table}{0}
\setcounter{algorithm}{0}
\renewcommand\thesection{\Alph{section}}
\renewcommand\thefigure{\Alph{figure}}
\renewcommand\thetable{\Alph{table}}
\renewcommand\thealgorithm{\Alph{algorithm}}

In the supplementary materials, we provide an algorithmic overview in Section \ref{sec:alg} and ablation study of our model in Section \ref{sec:ablation}. We give more quantitative and qualitative results in Section \ref{sec:result1} and Section \ref{sec:result2}. Finally, we give more implementation details in Section \ref{sec:implement}.

\section{Algorithmic Overview}
\label{sec:alg}

\begin{algorithm}[!ht]

	\caption{Distilling self-supervised knowledge by Embedding Graph Alignment using a {\em simultaneous} training strategy
	} 
	\label{Algorithm1}
	
	\parbox{3.25in}{
		\textbf{Input:} Training image data $\mathcal{I}$ with labels $Y$, Training iterations $\tau$; Self-supervised teacher networks $\mathcal{N}_t$. 
		\\ [0.05cm]
		\textbf{Output:} 
		Distilled student networks $\mathcal{N}_s$. 
		\\[0.05cm]
		\textbf{Initialisation:} 
		Randomly initialise $\mathcal{N}_s$, teacher's new added layers $\mathcal{M}_t$, two optimizers.
		\\ [0.05cm]
		\textbf{for} iteration $t$ in {[}1 : $\tau${]} : \\
		\hphantom{~~~~~~}
		{Feed a batch of $B$ images $\{ I_1, I_2, ..., I_B\}$ to $\mathcal{N}_t$, $\mathcal{N}_s$.}
		\\ [0.05cm]
		\hphantom{~~~~~~}
		Get teacher feature embeddings $X_t$.
		\\ [0.05cm]
		\hphantom{~~~~~~}
		Get student feature embeddings $X_s$. 
		\\ [0.05cm]
		\hphantom{~~~~~~}
		Derive teacher edge matrix  $E_t = E(X_t, X_t)$.
		\\ [0.05cm]
		\hphantom{~~~~~~} 
		Derive student edge matrix $E_s = E(X_s, X_s)$.
		\\ [0.05cm]
		\hphantom{~~~~~~}
		Derive node matrix $N_{st} = E (X_t, X_s)$.
		\\ [0.05cm]
		\\ [0.05cm]
		\hphantom{~~~~~~}
	    Compute edge matching loss $\mathcal{L}_{edge} = 
        {\left\|{{E}_t - {E}_s}\right\|_{2}}$.\\ [0.05cm]
		\hphantom{~~~~~~}
		Compute node matching loss $\mathcal{L}_{node} =
{\left\|{N_{st} - \mathcal{I}}\right\|_{2}}$.
        \\ [0.05cm]
		\hphantom{~~~~~~}
		Compute the EGA loss $\mathcal{L}_{EGA} = \mathcal{L}_{node} +\lambda \mathcal{L}_{edge}$.
		\\ [0.05cm]
		\hphantom{~~~~~~}
		Compute teacher final loss
		$\mathcal{L}_t  = \mathcal{L}_{ce_{t}}$.\\ [0.05cm]
		\hphantom{~~~~~~}
		Compute student final loss
		$\mathcal{L}_s  = \mathcal{L}_{ce_{s}} + \lambda_{EGA} \mathcal{L}_{EGA}$.
		\\ [0.05cm]
		\hphantom{~~~~~~}
		 Backpropagation on the teacher's new layers $\mathcal{M}_t$.\\ [0.05cm]
		\hphantom{~~~~~~}
		Backpropagation on the student network $\mathcal{N}_s$.
		\\[0.05cm]
\textbf{end for}
\\
\textbf{return $\mathcal{N}_t$}
	}
\end{algorithm}

As mentioned in the paper in Section 3.4, we adopt two different training strategies for optimizing the teacher and the student networks: (1) mutual learning \cite{zhang2018deep}, which optimizes the teacher with the target objective $\mathcal{L}_{ce}$ and the student with the full objective {\em simultaneously}; (2) sequential learning \cite{hinton2015distilling}, which optimizes the teacher and student networks {\em sequentially} by first training the teacher with the task objective $\mathcal{L}_{ce}$ and then training the student with the full objective. We present the algorithmic overviews for our Embedding Graph Alignment (EGA) model using these two training strategies in Algorithm \ref{Algorithm1} and Algorithm \ref{Algorithm2} respectively.

\begin{algorithm}[!ht]

	\caption{Distilling self-supervised knowledge by Embedding Graph Alignment using a {\em sequential} training strategy
	} 
	\label{Algorithm2}
	
	\parbox{3.25in}{
		\textbf{Input:} Training image data $\mathcal{I}$ with labels $Y$, Training iterations $\tau$; Self-supervised teacher networks $\mathcal{N}_t$. Pre-trained teacher's new added layers $\mathcal{M}_t$.
		\\ [0.05cm]
		\textbf{Output:} 
		Distilled student networks  $\mathcal{N}_s$. 
		\\[0.05cm]
		\textbf{Initialisation:} 
		Randomly initialise $\mathcal{N}_s$, student optimizer.
		\\ [0.05cm]
		\textbf{for} iteration $t$ in {[}1 : $\tau${]} : \\
		\hphantom{~~~~~~}
		{Feed a batch of $B$ images $\{ I_1, I_2, ..., I_B\}$ to $\mathcal{N}_t$, $\mathcal{N}_s$.}
		\\ [0.05cm]
		\hphantom{~~~~~~}
		Get teacher feature embeddings $X_t$.
		\\ [0.05cm]
		\hphantom{~~~~~~}
		Get student feature embeddings $X_s$. 
		\\ [0.05cm]
		\hphantom{~~~~~~}
		Derive teacher edge matrix  $E_t = E(X_t, X_t)$.
		\\ [0.05cm]
		\hphantom{~~~~~~} 
		Derive student edge matrix $E_s = E(X_s, X_s)$.
		\\ [0.05cm]
		\hphantom{~~~~~~}
		Derive node matrix $N_{st} = E (X_t, X_s)$.
		\\ [0.05cm]
		\\ [0.05cm]
		\hphantom{~~~~~~}
	    Compute edge matching loss $\mathcal{L}_{edge} = 
        {\left\|{{E}_t - {E}_s}\right\|_{2}}$.\\ [0.05cm]
		\hphantom{~~~~~~}
		Compute node matching loss $\mathcal{L}_{node} =
{\left\|{N_{st} - \mathcal{I}}\right\|_{2}}$.
        \\ [0.05cm]
		\hphantom{~~~~~~}
		Compute the EGA loss $\mathcal{L}_{EGA} = \mathcal{L}_{node} +\lambda \mathcal{L}_{edge}$.
		\\ [0.05cm]
		\hphantom{~~~~~~}
		Compute student final loss
		$\mathcal{L}_s  = \mathcal{L}_{ce_{s}} + \lambda_{EGA} \mathcal{L}_{EGA}$.
		\\ [0.05cm]
		\hphantom{~~~~~~}
		Backpropagation on the student network $\mathcal{N}_s$.
		\\[0.05cm]
\textbf{end for}
\\
\textbf{return $\mathcal{N}_t$}
	}
\end{algorithm}

\section{Ablation Study}
\label{sec:ablation}

\myparagraph{Model ablation.} We analyze the different loss terms in our EGA model. As Table \ref{tab:ablation} shows, we evaluate our model in comparison to three baseline models: (1) baseline (which is the student network trained without distillation); (2) EGA w/o $\mathcal{L}_{node}$, which removes the loss term $\mathcal{L}_{node}$ for aligning node matrix to an identity matrix; (3) EGA w/o $\mathcal{L}_{edge}$, which removes the loss term $\mathcal{L}_{edge}$ for aligning the edge matrices between the teacher graph and the student graph. From Table \ref{tab:ablation}, we have the following observations. First, our full model formulation EGA performs much better than the other EGA variants with a single loss term,  offering the best  accuracy of 76.65\%, 84.15\%, 60.61\% on CIFAR100, STL-10 and TinyImageNet. Second, ``EGA w/o $\mathcal{L}_{node}$'' and ``EGA w/o $\mathcal{L}_{edge}$'' both outperform the baseline, which suggests that both loss terms are effective for distilling the knowledge from the teacher to the student network. This ablation shows that our EGA works effectively for distilling self-supervised knowledge to improve recognition. 

{
\begin{table}[!ht]
    \centering 
	\scalebox{0.8}{
	\begin{tabular}{l|c|c|c}
		 \hline
		 Method & CIFAR 100 & STL-10 & TinyImageNet \\
		 \hline \hline 
		 baseline & 
		72.38 & 80.63 & 57.54
		\\ \hline

		EGA w/o $\mathcal{L}_{node}$ &
		75.18 & 83.78 & 59.52 
		\\ 
		EGA w/o $\mathcal{L}_{edge}$ &
		75.21 & 82.35 & 59.82 
		\\ 
		\bf EGA & 
		\bf 76.65 & \bf 84.15 & \bf 60.61 
		\\
		\hline 
	\end{tabular}
    }
    \vskip 1em
	\caption{
	Ablation study on different loss components. 
	Teacher: self-supervised ViT-B/32 from \cite{radford2021learning}. 
	Student: resnet8x4.
	} 
	\label{tab:ablation}
    \vskip -1.5em
\end{table}
}

\myparagraph{Hyperparameter analysis.} 
We analyze the hyperparameters of the edge matching loss $\mathcal{L}_{edge}$ and node matching loss $\mathcal{L}_{node}$ in Table \ref{tab:hyper}. We explore the effect of these two loss terms by varying their weights $\lambda_{edge}$ and $\lambda_{node}$ wrt to the supervised loss term (i.e. $\mathcal{L}  = \mathcal{L}_{ce} + \lambda_{edge} \mathcal{L}_{edge}$ and $\mathcal{L}  = \mathcal{L}_{ce} + \lambda_{node} \mathcal{L}_{node}$). We test this by setting the weight of one loss term to be zero and changing the weight of another loss term (either $\lambda_{edge}$ or $\lambda_{node}$). 
In the first two rows of Table \ref{tab:hyper}, we can see that the model performance first increases and then decreases when increasing $\lambda_{edge}$. In the last two rows of the table, 
We can also see that similar trends of model performance as the left figure when increasing $\lambda_{node}$. The best performing values for $\lambda_{edge}$, $\lambda_{node}$ are 0.5, 1.5, which we use to trade off $\mathcal{L}_{node}$, $\mathcal{L}_{edge}$ and guide our hyperpameter setting $\lambda=\frac{\lambda_{edge}}{\lambda_{node}}\approx0.3$.

{
\setlength{\tabcolsep}{2pt}
\renewcommand{\arraystretch}{1.1}
\begin{table}[!ht]
    \centering 
	\scalebox{0.8}{
	\begin{tabular}{l|c|c|c|c|c|c}
        \hline 
		 Node weight & 1.2 & 1.4 & 1.5 & 1.6 & 1.8 & 2.0 \\
		 \hline
		 Accuracy & 
		75.11 & 75.56 & \bf 75.85 & 75.17 & 75.51 & 74.91 \\ 
		 \hline 
		 \hline 
	     Edge weight & 0.2 & 0.4 & 0.5 & 0.6 & 0.8 & 1.0 \\
		 \hline
		 Accuracy & 
		74.47 & 75.32 & \bf 75.56 & 74.99 & 74.72 & 74.27 \\
		\hline 
	\end{tabular}
	}
	\vskip 1em
	\caption{
	Effect of hyperparameters (node weight, edge weight) on CIFAR10.
	} 
	\label{tab:hyper}
	\vskip -0.5em
\end{table}
}

\section{More Quantitative Results}
\label{sec:result1}

\begin{figure}[!ht]
    \centering
        \includegraphics[width=0.4\linewidth]{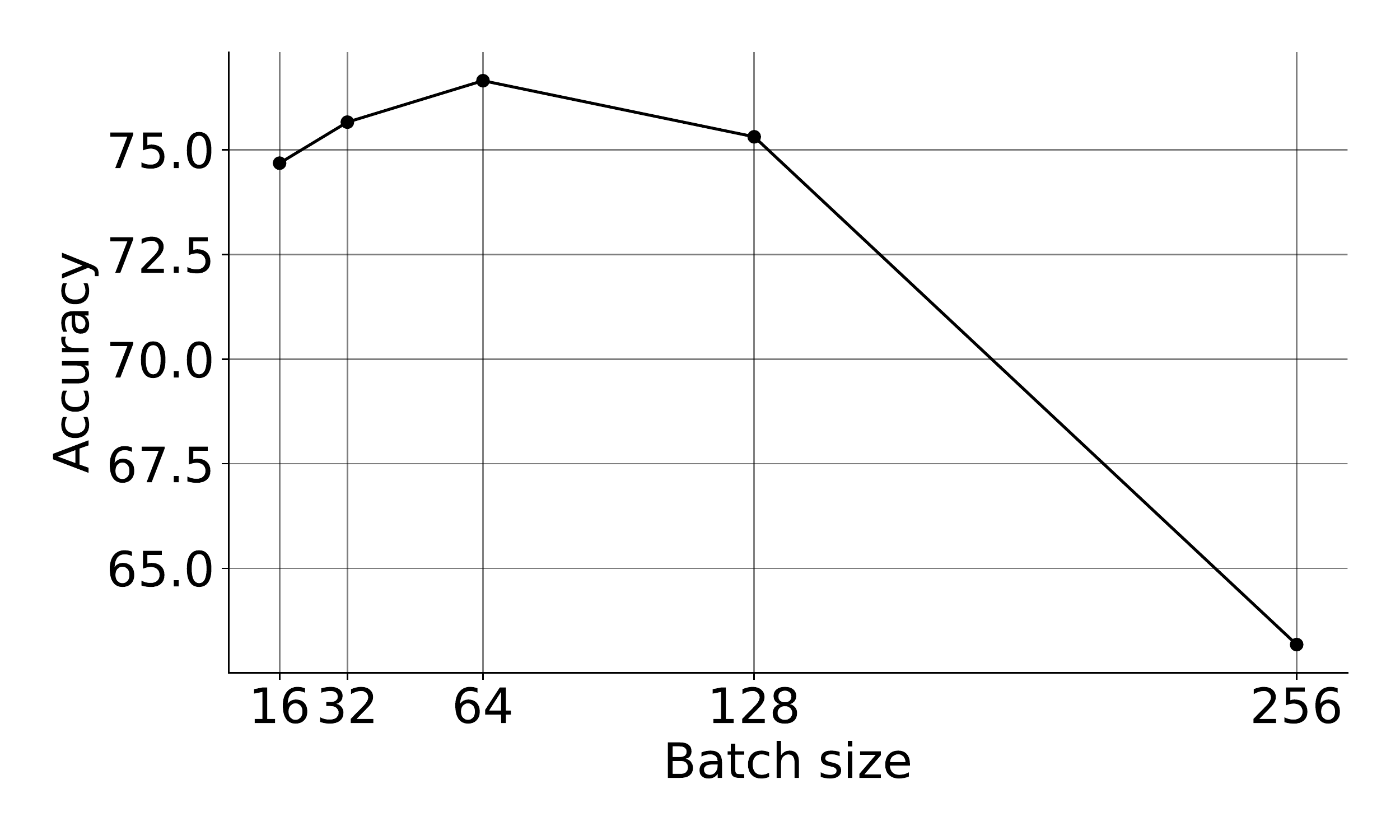}
    \vskip 0.5em
    \caption{Model performance (top 1 accuracy, \%) under varying graph size (i.e. which is quantified by the batch size).}
\label{fig:graphsize}
\end{figure}

\myparagraph{Analysis of the graph size.} 
We analyze the different graph size of the EGA loss in Figure \ref{fig:graphsize}. As the graph size (i.e. the number of node) is equal to the batch size, we change the batch size from 16, 32, 64, 128, 256 with a double increase to obtain different graph size. Figure \ref{fig:graphsize} shows the trends of model performance as the graph size change. As can be seen, EGA obtains the best model performance with a graph size of 64, and the worse performance with a graph size of 256. When using a graph size of 16, 32, 64, and 128, the model performs similarly well, which suggests that EGA could work robustly under various choices of graph size. 

{
\setlength{\tabcolsep}{2pt}
\renewcommand{\arraystretch}{1.1}
\begin{table*}[!ht]
    \centering 
	\scalebox{0.8}{
	\begin{tabular}{l|c|c|c|c|c|c}
	\hline \multirow{2}{10pt}{Method} 
		& \multicolumn{3}{c|}{Same student different teacher} 
		& \multicolumn{3}{c}{Same teacher different student} \\ 
		\cline{2-4}	\cline{5-7}   & ViT-B/32 & ViT-B/16 & RN101 &Resnet8x4 & ShuffleNetV1 & VGG13
	\\	\hline	\hline
		KD\cite{hinton2015distilling}  & 
		71.82 & 9.6 & 65.16 &71.82 & 73.52 & 75.40
       \\ 
        FitNet\cite{romero2014fitnets} & 
		  74.83 & 67.85 & 73.69 & 74.83 & nan & 76.28
		\\

		PKT \cite{passalis2018learning} & 
		  73.48 & 72.71 & 72.66 & 73.48 & 74.92 & 75.35
		\\
		RKD\cite{park2019relational} & 
         73.36 &  72.43 & 73.92 &73.36 & 72.62 & 73.26 
		  \\
		NCE \cite{chen2021distilling} & 
		  74.42 & 73.89 & 74.04 & 74.42 & 74.36 & 76.61
		\\
		CRD\cite{tian2019contrastive} \quad & 
		 75.51 & 73.38 & 74.85  & 75.51 & 74.87 &77.41
		\\
		CCL\cite{chen2021distilling} \quad & 
		 75.98 & 39.56 & 74.22 & 75.98 & 76.05 &77.54 
        \\
		\hline\hline
		\bf EGA \quad & 
		\bf  76.11 & 
		\bf 74.02 &
		\bf  75.22 &
		\bf  76.11 &
		\bf  76.74 &
		\bf  77.76 \\
		\hline
	\end{tabular}
	}
    \vskip 0.5em
	\caption{
	Top 1 accuracy (\%) of student networks on CIFAR100. 
	In column 1-3, we use the same student (resnet8x4) and vary the self-supervised teacher using ViT-B/32, ViT-B/16, RN101 from \cite{radford2021learning}. 
	In column 4-6,  we use the same teacher (ViT-B/32) and vary the student using resnet8x4, shuffleNetV1, VGG13. 
	The teacher and student 
	are trained {\bf \em sequentially} \cite{zhang2018deep}. 
	Best results are {\bf bold}.
	} 
	\label{tab:network2}
	\vskip -0.5em
\end{table*}
}
{
\begin{table}[!ht]
    \centering 
	\scalebox{0.8}{
	\begin{tabular}{l|c|c|c}
        \hline 
	     Method & CIFAR 100 & STL-10 & TinyImageNet \\
		 \hline \hline 
		KD \cite{hinton2015distilling} &
		71.82 & 82.51 & 53.88
		\\
        FitNet & 
       74.83  & nan &  nan
        \\
        PKT \cite{passalis2018learning} & 
		 73.48  & nan & 59.72
		\\
		RKD \cite{park2019relational}& 
		73.36 & 82.67 & 58.32
		\\ 
		NCE \cite{chen2021distilling} & 
		 74.42 & 80.07 & 60.00
		\\
		CRD \cite{tian2019contrastive}& 
		75.51 & 78.76 & 60.82 
		\\ 
        CCL\cite{chen2021distilling} &
        75.98 & 80.41 & 61.24 
		\\
		\hline \hline
		\bf EGA & 
		\bf 76.11 & \bf 83.01 & \bf 61.85
		\\
		\hline 
	\end{tabular}
	}
    \vskip 0.5em
	\caption{
	Top 1 accuracy (\%) of student networks on CIFAR100, STL10, TinyImageNet. 
	Teacher: self-supervised ViT-32 \cite{radford2021learning}. 
	Student: resnet8x4. 
    The teacher and student are trained 
	{\bf \em sequentially} \cite{hinton2015distilling}. 
	Best results are highlighted in {\bf bold}. 
	} 
	\label{tab:data2}
	\vskip -0.5em
\end{table}
}
\myparagraph{Numerical results.} Due to space limit, we show only the top performing methods in Table 2 and Table 5 in the paper. Here, we present the numerical results of all the comparative methods using the same evaluation setup in Table 2 and Table 5, as detailed in Table \ref{tab:network2} and Table \ref{tab:data2} respectively.
As can be seen, our model obtains the best overall performance compared to the other compared methods. These evidences are consistent with our analysis in the paper. 

\section{More Qualitative Results}
\label{sec:result2}

\begin{figure}[!ht]
    \centering
        \includegraphics[width=0.3\linewidth]{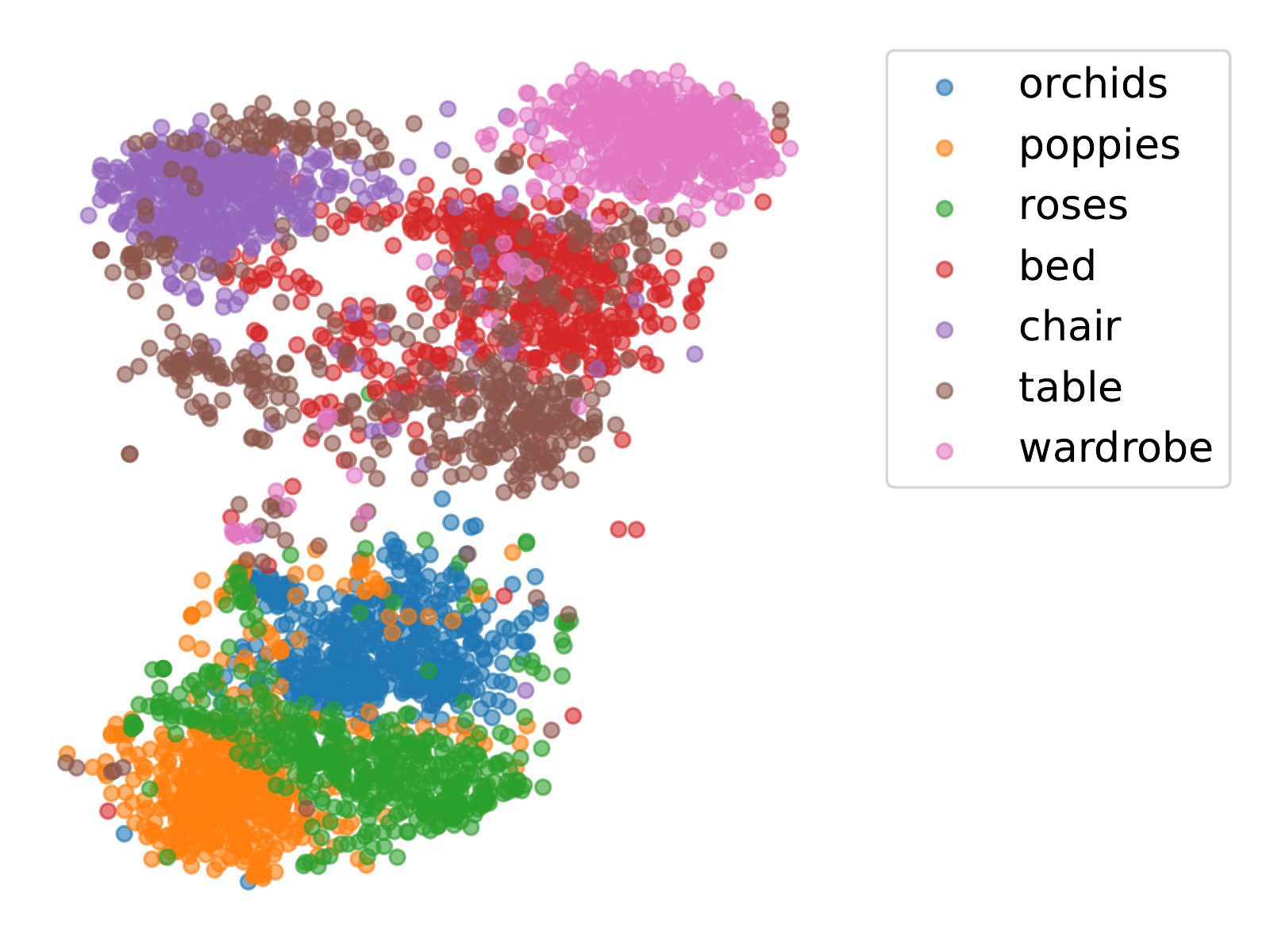}
    \vskip 0.5em
    \caption{t-SNE visualization \cite{maaten2008visualizing} of the RKD feature embeddings from different classes on CIFAR100.}
    \vskip -0.5em
\label{fig:tsneRKD}
\end{figure}

\begin{figure}[htbp]
    \centering
        \includegraphics[width=0.3\linewidth]{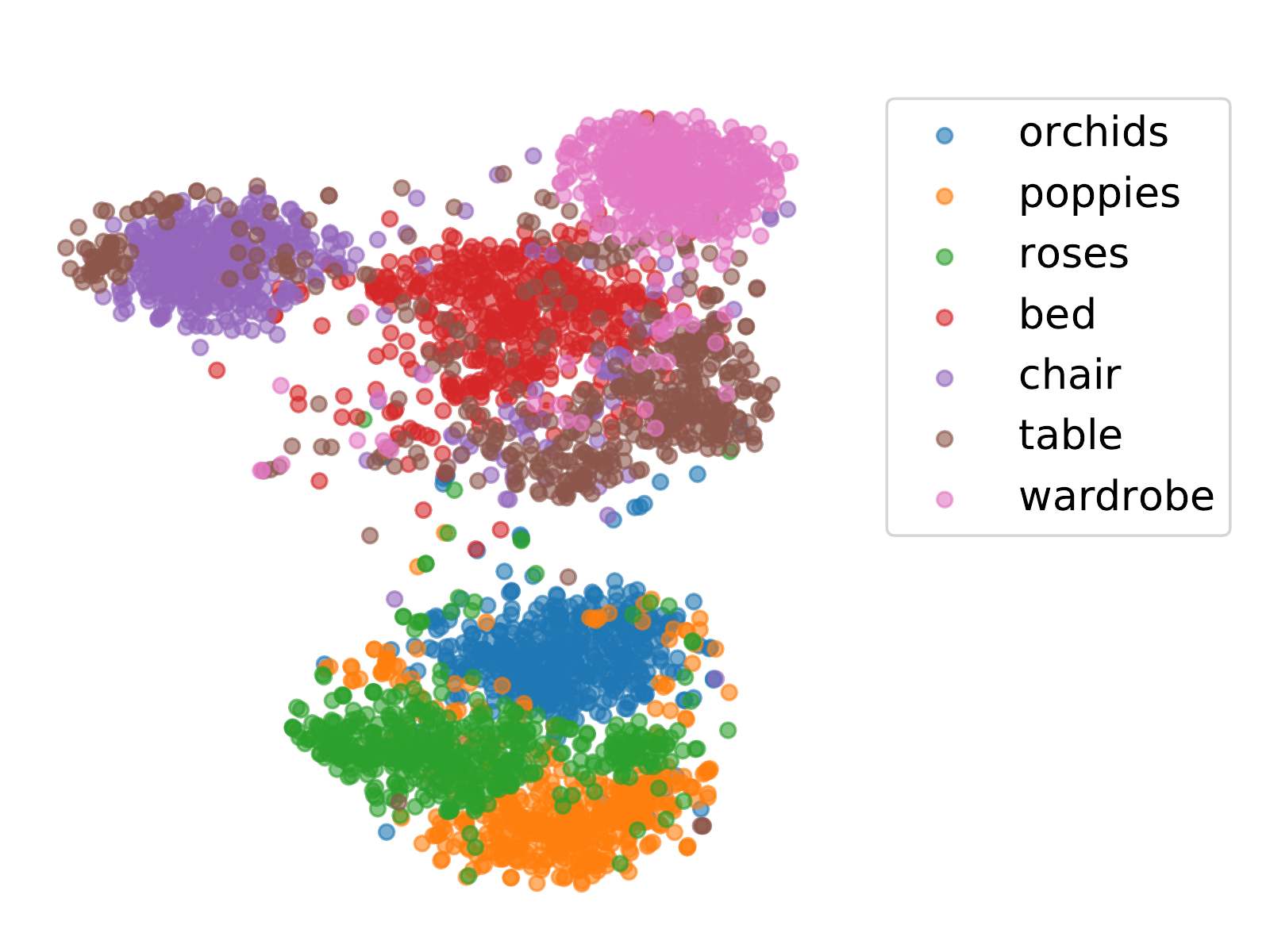}
    \vskip 0.5em
    \caption{t-SNE visualization \cite{maaten2008visualizing} of the EGA feature embeddings from different classes on CIFAR100.}
    \vskip -0.5em
\label{fig:tsneGM}
\end{figure}

\myparagraph{Visualizing embeddings with t-SNE.}
In order to examine the distilled structural semantic relationships learned from the teacher network by different loss terms, we visualize the feature embeddings of the EGA and RDK models from different classes. We compare RKD and our EGA, as both of these models focus on distilling structural relationship information from the teacher networks. 
Figure \ref{fig:tsneRKD} and Figure \ref{fig:tsneGM} 
show the feature distributions from RKD and EGA models, which depicts the seven fine-grained classes from two hyperclasses -- {\em orchids}, {\em poppies}, {\em roses} belong to the hyperclass {\em flowers}, while {\em bed}, {\em chair}, {\em table}, and {\em wardrobe} belong to the hyperclass {\em furniture}. 
As can be seen in two figures, our model is capable of capturing the inter-class and hyper-class structural relations much better than RKD. 
From Figure \ref{fig:tsneRKD}, We can see that the cluster of {\em bed} is mixed with other clusters, especially the cluster of {\em table}. While in Figure \ref{fig:tsneGM}, the cluster of {\em bed} has higher compactness, and {\em table} has clearer decision boundary wrt the clusters of other classes than RKD's. 
Moreover, the hyperclasses {\em furniture} and {\em flowers} are more separated in our model EGA than RKD. Overall, these visualizations suggest that the EGA model learns the structural semantics and captures the instance-instance relations better than RKD.


\section{Implementation Details}
\label{sec:implement}

\myparagraph{Data augmentation.}
Each input image is transformed into two ways for teacher and student model.
For student networks, The image augmentation pipeline consists of the following transformations: random cropping, random horizontal flipping, and normalization. 
As for teacher, the pipeline consists of random cropping, resizing to 224 × 224, random horizontal flipping and normalization. 
If the teacher model is a self-supervised pre-trained model which has its own preprocessing parameter, e.g. CLIP \cite{radford2021learning}, both our teacher and our student will follow the same CLIP image preprocessing normalization parameters. The normalization parameters for evaluation is always consistent with training.

\myparagraph{Training Details.}
The student networks are trained by the combination of a cross-entropy classification objective and a knowledge distillation objective. For the weight balance factor of these two loss terms, we directly use the optimal value from the original paper or adopted from CRD \cite{tian2019contrastive}.
For the {\em simultaneous} training mode, the teacher's new added layers are trained by SGD with an initial learning rate of 0.01. For student network, we follow the setting from CRD \cite{tian2019contrastive}. That is, if the student is ShuffleNetV1, we use an initial learning rate of 0.01, while 0.05 for other models.

\myparagraph{Supervised teacher vs self-supervised teacher networks.} As mentioned in our paper, we evaluate multiple self-supervised teacher networks including ViT-B/32, ViT-B/16, RN101 from CLIP \cite{radford2021learning}, which are downloaded from the \href{https://github.com/openai/CLIP}{CLIP GitHub repository}. We only fine-tune the last two layers of the teacher networks on the target datasets. Thus, a self-supervised teacher network would capture both self-supervised knowledge and supervised knowledge from a target dataset. 
As for the supervised teacher networks, we use the pre-trained teacher networks 
which are pre-trained on ImageNet \cite{deng2009imagenet} and fine-tuned on the CIFAR100. Hence, a supervised teacher network would contain the supervised knowledge from the ImageNet and CIFAR100, which requires more annotations efforts compared to a self-supervised teacher network. For instance, in the paper, the self-supervised RN101 teacher network in Table 2 obtains an accuracy of 67.76\% while the supervised RN101 teacher network in Table 3 obtains an accuracy of 73.58\%. As for the downstream task on CIFAR100, the student with a self-supervised RN101 teacher obtains an accuracy of 75.22\% (in Table 2) while the student with a supervised RN101 teacher obtains an accuracy of 75.77\% (in Table 3). Overall, these results suggest that a self-supervised teacher network may not start with a better model initialization compared to a supervised teacher; however, with our model formulation, the self-supervised knowledge can be well distilled to improve the generalization of a student network on the downstream tasks. 

\bibliography{reference}

\begin{thebibliography}{42}
\providecommand{\natexlab}[1]{#1}
\providecommand{\url}[1]{\texttt{#1}}
\expandafter\ifx\csname urlstyle\endcsname\relax
  \providecommand{\doi}[1]{doi: #1}\else
  \providecommand{\doi}{doi: \begingroup \urlstyle{rm}\Url}\fi

\bibitem[Afouras et~al.(2020)Afouras, Chung, and Zisserman]{afouras2020asr}
Triantafyllos Afouras, Joon~Son Chung, and Andrew Zisserman.
\newblock Asr is all you need: Cross-modal distillation for lip reading.
\newblock In \emph{ICASSP}, 2020.

\bibitem[Albanie et~al.(2018)Albanie, Nagrani, Vedaldi, and
  Zisserman]{albanie2018emotion}
Samuel Albanie, Arsha Nagrani, Andrea Vedaldi, and Andrew Zisserman.
\newblock Emotion recognition in speech using cross-modal transfer in the wild.
\newblock In \emph{ACM MM}, 2018.

\bibitem[Aytar et~al.(2016)Aytar, Vondrick, and Torralba]{aytar2016soundnet}
Yusuf Aytar, Carl Vondrick, and Antonio Torralba.
\newblock Soundnet: Learning sound representations from unlabeled video.
\newblock In \emph{NeuRIPS}, 2016.

\bibitem[Ba and Caruana(2014)]{ba2014deep}
Jimmy Ba and Rich Caruana.
\newblock Do deep nets really need to be deep?
\newblock In \emph{NeuRIPS}, 2014.

\bibitem[Bachman et~al.(2019)Bachman, Hjelm, and
  Buchwalter]{bachman2019learning}
Philip Bachman, R~Devon Hjelm, and William Buchwalter.
\newblock Learning representations by maximizing mutual information across
  views.
\newblock In \emph{NeuRIPS}, 2019.

\bibitem[Buciluǎ et~al.(2006)Buciluǎ, Caruana, and
  Niculescu-Mizil]{bucilua2006model}
Cristian Buciluǎ, Rich Caruana, and Alexandru Niculescu-Mizil.
\newblock Model compression.
\newblock In \emph{ACM SIGKDD}, 2006.

\bibitem[Chen et~al.(2020{\natexlab{a}})Chen, Kornblith, Norouzi, and
  Hinton]{chen2020simple}
Ting Chen, Simon Kornblith, Mohammad Norouzi, and Geoffrey Hinton.
\newblock A simple framework for contrastive learning of visual
  representations.
\newblock In \emph{ICML}, 2020{\natexlab{a}}.

\bibitem[Chen et~al.(2020{\natexlab{b}})Chen, Kornblith, Swersky, Norouzi, and
  Hinton]{chen2020big}
Ting Chen, Simon Kornblith, Kevin Swersky, Mohammad Norouzi, and Geoffrey~E
  Hinton.
\newblock Big self-supervised models are strong semi-supervised learners.
\newblock In \emph{NeuRIPS}, 2020{\natexlab{b}}.

\bibitem[Chen and He(2021)]{chen2020exploring}
Xinlei Chen and Kaiming He.
\newblock Exploring simple siamese representation learning.
\newblock In \emph{CVPR}, 2021.

\bibitem[Chen et~al.(2021)Chen, Xian, Koepke, Shan, and
  Akata]{chen2021distilling}
Yanbei Chen, Yongqin Xian, A~Koepke, Ying Shan, and Zeynep Akata.
\newblock Distilling audio-visual knowledge by compositional contrastive
  learning.
\newblock In \emph{CVPR}, 2021.

\bibitem[Deng et~al.(2009)Deng, Dong, Socher, Li, Li, and
  Fei-Fei]{deng2009imagenet}
Jia Deng, Wei Dong, Richard Socher, Li-Jia Li, Kai Li, and Li~Fei-Fei.
\newblock Imagenet: A large-scale hierarchical image database.
\newblock In \emph{CVPR}, 2009.

\bibitem[Doersch et~al.(2015)Doersch, Gupta, and
  Efros]{doersch2015unsupervised}
Carl Doersch, Abhinav Gupta, and Alexei~A Efros.
\newblock Unsupervised visual representation learning by context prediction.
\newblock In \emph{ICCV}, 2015.

\bibitem[Gidaris et~al.(2018)Gidaris, Singh, and
  Komodakis]{gidaris2018unsupervised}
Spyros Gidaris, Praveer Singh, and Nikos Komodakis.
\newblock Unsupervised representation learning by predicting image rotations.
\newblock In \emph{ICLR}, 2018.

\bibitem[Gupta et~al.(2016)Gupta, Hoffman, and Malik]{gupta2016cross}
Saurabh Gupta, Judy Hoffman, and Jitendra Malik.
\newblock Cross modal distillation for supervision transfer.
\newblock In \emph{CVPR}, 2016.

\bibitem[Hadsell et~al.(2006)Hadsell, Chopra, and
  LeCun]{hadsell2006dimensionality}
Raia Hadsell, Sumit Chopra, and Yann LeCun.
\newblock Dimensionality reduction by learning an invariant mapping.
\newblock In \emph{CVPR}, 2006.

\bibitem[He et~al.(2020)He, Fan, Wu, Xie, and Girshick]{he2019momentum}
Kaiming He, Haoqi Fan, Yuxin Wu, Saining Xie, and Ross Girshick.
\newblock Momentum contrast for unsupervised visual representation learning.
\newblock In \emph{CVPR}, 2020.

\bibitem[Hinton et~al.(2015)Hinton, Vinyals, and Dean]{hinton2015distilling}
Geoffrey Hinton, Oriol Vinyals, and Jeff Dean.
\newblock Distilling the knowledge in a neural network.
\newblock \emph{arXiv preprint arXiv:1503.02531}, 2015.

\bibitem[Hjelm et~al.(2019)Hjelm, Fedorov, Lavoie-Marchildon, Grewal, Bachman,
  Trischler, and Bengio]{hjelm2018learning}
R~Devon Hjelm, Alex Fedorov, Samuel Lavoie-Marchildon, Karan Grewal, Phil
  Bachman, Adam Trischler, and Yoshua Bengio.
\newblock Learning deep representations by mutual information estimation and
  maximization.
\newblock In \emph{ICLR}, 2019.

\bibitem[Jia et~al.(2021)Jia, Yang, Xia, Chen, Parekh, Pham, Le, Sung, Li, and
  Duerig]{jia2021scaling}
Chao Jia, Yinfei Yang, Ye~Xia, Yi-Ting Chen, Zarana Parekh, Hieu Pham, Quoc~V
  Le, Yunhsuan Sung, Zhen Li, and Tom Duerig.
\newblock Scaling up visual and vision-language representation learning with
  noisy text supervision.
\newblock In \emph{ICML}, 2021.

\bibitem[Larsson et~al.(2017)Larsson, Maire, and
  Shakhnarovich]{larsson2017colorization}
Gustav Larsson, Michael Maire, and Gregory Shakhnarovich.
\newblock Colorization as a proxy task for visual understanding.
\newblock In \emph{CVPR}, 2017.

\bibitem[Liu et~al.(2019)Liu, Cao, Li, Yuan, Hu, Li, and
  Duan]{liu2019knowledge}
Yufan Liu, Jiajiong Cao, Bing Li, Chunfeng Yuan, Weiming Hu, Yangxi Li, and
  Yunqiang Duan.
\newblock Knowledge distillation via instance relationship graph.
\newblock In \emph{CVPR}, 2019.

\bibitem[Maaten and Hinton(2008)]{maaten2008visualizing}
Laurens van~der Maaten and Geoffrey Hinton.
\newblock Visualizing data using t-sne.
\newblock \emph{JMLR}, 2008.

\bibitem[Misra and Maaten(2020)]{misra2020self}
Ishan Misra and Laurens van~der Maaten.
\newblock Self-supervised learning of pretext-invariant representations.
\newblock In \emph{CVPR}, 2020.

\bibitem[Noroozi and Favaro(2016)]{noroozi2016unsupervised}
Mehdi Noroozi and Paolo Favaro.
\newblock Unsupervised learning of visual representations by solving jigsaw
  puzzles.
\newblock In \emph{ECCV}, 2016.

\bibitem[Noroozi et~al.(2017)Noroozi, Pirsiavash, and
  Favaro]{noroozi2017representation}
Mehdi Noroozi, Hamed Pirsiavash, and Paolo Favaro.
\newblock Representation learning by learning to count.
\newblock In \emph{ICCV}, 2017.

\bibitem[Novotny et~al.(2018)Novotny, Albanie, Larlus, and
  Vedaldi]{novotny2018self}
David Novotny, Samuel Albanie, Diane Larlus, and Andrea Vedaldi.
\newblock Self-supervised learning of geometrically stable features through
  probabilistic introspection.
\newblock In \emph{CVPR}, 2018.

\bibitem[Oord et~al.(2018)Oord, Li, and Vinyals]{oord2018representation}
Aaron van~den Oord, Yazhe Li, and Oriol Vinyals.
\newblock Representation learning with contrastive predictive coding.
\newblock \emph{arXiv preprint arXiv:1807.03748}, 2018.

\bibitem[Park et~al.(2019)Park, Kim, Lu, and Cho]{park2019relational}
Wonpyo Park, Dongju Kim, Yan Lu, and Minsu Cho.
\newblock Relational knowledge distillation.
\newblock In \emph{CVPR}, 2019.

\bibitem[Passalis and Tefas(2018)]{passalis2018learning}
Nikolaos Passalis and Anastasios Tefas.
\newblock Learning deep representations with probabilistic knowledge transfer.
\newblock In \emph{ECCV}, 2018.

\bibitem[Passalis et~al.(2020)Passalis, Tzelepi, and
  Tefas]{passalis2020heterogeneous}
Nikolaos Passalis, Maria Tzelepi, and Anastasios Tefas.
\newblock Heterogeneous knowledge distillation using information flow modeling.
\newblock In \emph{CVPR}, 2020.

\bibitem[Radford et~al.(2021)Radford, Kim, Hallacy, Ramesh, Goh, Agarwal,
  Sastry, Askell, Mishkin, Clark, et~al.]{radford2021learning}
Alec Radford, Jong~Wook Kim, Chris Hallacy, Aditya Ramesh, Gabriel Goh,
  Sandhini Agarwal, Girish Sastry, Amanda Askell, Pamela Mishkin, Jack Clark,
  et~al.
\newblock Learning transferable visual models from natural language
  supervision.
\newblock In \emph{ICML}, 2021.

\bibitem[Romero et~al.(2015)Romero, Ballas, Kahou, Chassang, Gatta, and
  Bengio]{romero2014fitnets}
Adriana Romero, Nicolas Ballas, Samira~Ebrahimi Kahou, Antoine Chassang, Carlo
  Gatta, and Yoshua Bengio.
\newblock Fitnets: Hints for thin deep nets.
\newblock In \emph{ICLR}, 2015.

\bibitem[Santa~Cruz et~al.(2018)Santa~Cruz, Fernando, Cherian, and
  Gould]{santa2018visual}
Rodrigo Santa~Cruz, Basura Fernando, Anoop Cherian, and Stephen Gould.
\newblock Visual permutation learning.
\newblock \emph{IEEE TPAMI}, 2018.

\bibitem[Tian et~al.(2019)Tian, Krishnan, and Isola]{tian2019contrastive}
Yonglong Tian, Dilip Krishnan, and Phillip Isola.
\newblock Contrastive representation distillation.
\newblock In \emph{ICLR}, 2019.

\bibitem[Tian et~al.(2020)Tian, Sun, Poole, Krishnan, Schmid, and
  Isola]{tian2020makes}
Yonglong Tian, Chen Sun, Ben Poole, Dilip Krishnan, Cordelia Schmid, and
  Phillip Isola.
\newblock What makes for good views for contrastive learning.
\newblock In \emph{NeuRIPS}, 2020.

\bibitem[Tung and Mori(2019)]{tung2019similarity}
Frederick Tung and Greg Mori.
\newblock Similarity-preserving knowledge distillation.
\newblock In \emph{ICCV}, 2019.

\bibitem[Wu et~al.(2018)Wu, Xiong, Yu, and Lin]{wu2018unsupervised}
Zhirong Wu, Yuanjun Xiong, Stella~X Yu, and Dahua Lin.
\newblock Unsupervised feature learning via non-parametric instance
  discrimination.
\newblock In \emph{CVPR}, 2018.

\bibitem[Zagoruyko and Komodakis(2017)]{zagoruyko2016paying}
Sergey Zagoruyko and Nikos Komodakis.
\newblock Paying more attention to attention: Improving the performance of
  convolutional neural networks via attention transfer.
\newblock In \emph{ICLR}, 2017.

\bibitem[Zbontar et~al.(2021)Zbontar, Jing, Misra, LeCun, and
  Deny]{zbontar2021barlow}
Jure Zbontar, Li~Jing, Ishan Misra, Yann LeCun, and St{\'e}phane Deny.
\newblock Barlow twins: Self-supervised learning via redundancy reduction.
\newblock In \emph{ICML}, 2021.

\bibitem[Zhang et~al.(2016)Zhang, Isola, and Efros]{zhang2016colorful}
Richard Zhang, Phillip Isola, and Alexei~A Efros.
\newblock Colorful image colorization.
\newblock In \emph{ECCV}, 2016.

\bibitem[Zhang et~al.(2017)Zhang, Isola, and Efros]{zhang2017split}
Richard Zhang, Phillip Isola, and Alexei~A Efros.
\newblock Split-brain autoencoders: Unsupervised learning by cross-channel
  prediction.
\newblock In \emph{CVPR}, 2017.

\bibitem[Zhang et~al.(2018)Zhang, Xiang, Hospedales, and Lu]{zhang2018deep}
Ying Zhang, Tao Xiang, Timothy~M Hospedales, and Huchuan Lu.
\newblock Deep mutual learning.
\newblock In \emph{CVPR}, 2018.

\end{thebibliography}
\end{document}